\documentclass[final]{cvpr}

\usepackage{times}
\usepackage{epsfig}
\usepackage{graphicx}
\usepackage{amsmath}
\usepackage{amssymb}

\usepackage{color}
\usepackage[ruled,linesnumbered]{algorithm2e}
\usepackage{booktabs}
\usepackage{bigstrut}
\usepackage{multirow}
\usepackage[switch]{lineno}
\usepackage{amsopn}
\usepackage{comment}
\usepackage{algorithmic}
\usepackage{tabularx}
\usepackage{mathrsfs}
\usepackage{makecell}
\usepackage{subfigure,tabulary}
\usepackage[cmintegrals]{newtxmath}
\usepackage{enumitem}

\usepackage[pagebackref=false,breaklinks=true,colorlinks,bookmarks=false]{hyperref}

\def\cvprPaperID{2} 
\def\confYear{CVPRW 2021}

\newcolumntype{L}[1]{>{\raggedright\arraybackslash}p{#1}}
\newcolumntype{C}[1]{>{\centering\arraybackslash}p{#1}}
\newcolumntype{R}[1]{>{\raggedleft\arraybackslash}p{#1}}

\usepackage{pifont}
\newcommand{\cmark}{\ding{51}}%
\begin{document}

\title{End-to-End Interactive Prediction and Planning with Optical Flow Distillation for Autonomous Driving}

\author{
    Hengli Wang$^1$, Peide Cai$^1$, Rui Fan$^2$, Yuxiang Sun$^3$, and Ming Liu$^1$\\
    $^1$ The Hong Kong University of Science and Technology
    \\
    $^2$ University of California San Diego
    \\
    $^3$ The Hong Kong Polytechnic University\\
    {\tt\small \{hwangdf, pcaiaa\}@connect.ust.hk, rfan@ucsd.edu, sun.yuxiang@outlook.com, eelium@ust.hk}
}

\maketitle
\pagestyle{empty}
\thispagestyle{empty}

\begin{abstract}
With the recent advancement of deep learning technology, data-driven approaches for autonomous car prediction and planning have achieved extraordinary performance. Nevertheless, most of these approaches follow a non-interactive prediction and planning paradigm, hypothesizing that a vehicle's behaviors do not affect others. The approaches based on such a non-interactive philosophy typically perform acceptably in sparse traffic scenarios but can easily fail in dense traffic scenarios. Therefore, we propose an end-to-end interactive neural motion planner (INMP) for autonomous driving in this paper. Given a set of past surrounding-view images and a high definition map, our INMP first generates a feature map in bird's-eye-view space, which is then processed to detect other agents and perform interactive prediction and planning jointly. Also, we adopt an optical flow distillation paradigm, which can effectively improve the network performance while still maintaining its real-time inference speed. Extensive experiments on the nuScenes dataset and in the closed-loop Carla simulation environment demonstrate the effectiveness and efficiency of our INMP for the detection, prediction, and planning tasks. Our project page is at \url{sites.google.com/view/inmp-ofd}.
\end{abstract}

\section{Introduction}
\label{sec.introduction}

\begin{figure}[t]
    \centering
    \includegraphics[width=0.99\linewidth]{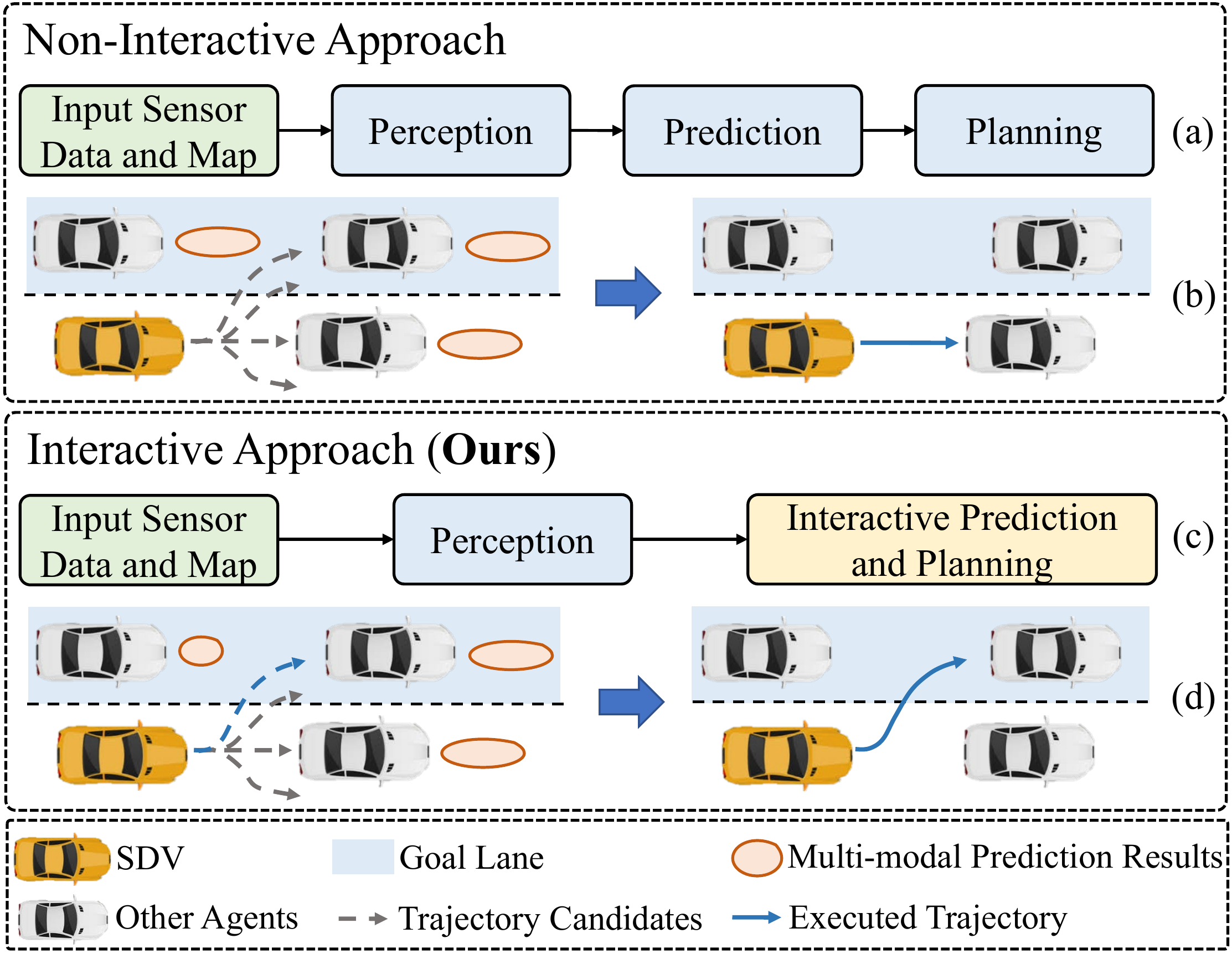}
    \caption{An illustration of non-interactive and our proposed interactive approaches, where (a) and (c) show the corresponding frameworks; and (b) and (d) show the corresponding driving performance in a dense traffic scenario. Specifically, the non-interactive SDV can struggle merging into the left lane, while our interactive SDV can perform a satisfactory lane merge by reasoning about how other agents will react to its behaviors.}
    \label{fig.demo}
\end{figure}

\begin{figure*}[t]
    \centering
    \includegraphics[width=0.99\textwidth]{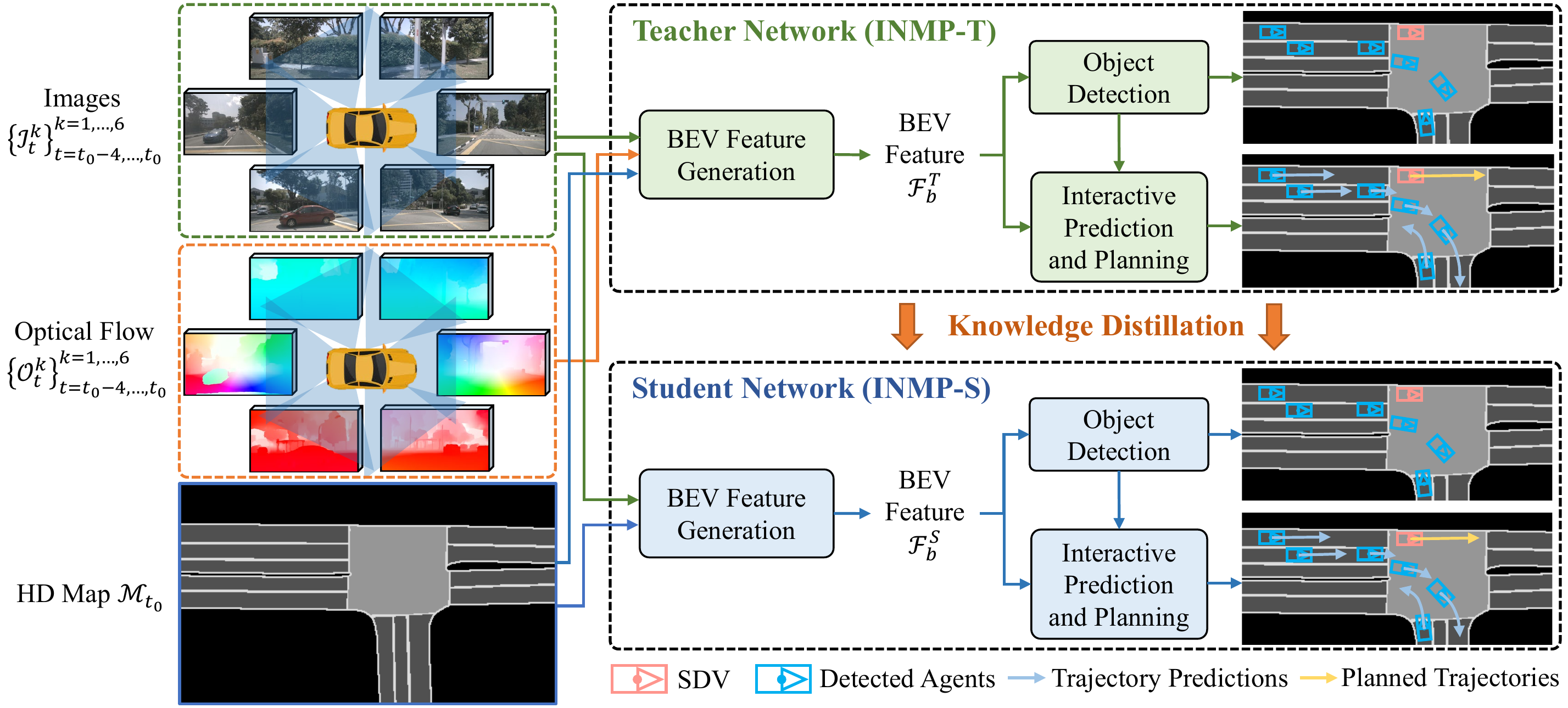}
    \caption{An overview of our INMP, which takes a set of past surrounding-view images and an HD map as input to jointly 1) detect other agents and 2) perform interactive prediction and planning. We also adopt an optical flow distillation paradigm, where the teacher network adopts a similar architecture to the student network but takes optical flow as an additional type of input. We then distill the knowledge from the teacher network to the student network, which can effectively improve the student network performance while still maintaining its real-time inference speed.}
    \label{fig.framework}
\end{figure*}

Autonomous driving aims at safely and efficiently maneuvering self-driving vehicles (SDVs) from a starting point to a target point with the input of sensor data and a pre-built map \cite{gao2017intention,codevilla2018end,cai2020vtgnet,liu2021hercules}. Most existing approaches are designed to follow a ``perception-prediction-planning'' paradigm, as shown in Fig.~\ref{fig.demo}~(a), where the perception module detects other agents from the sensor data, the prediction module estimates possible future trajectories of the detected agents, and the planning module generates a safe trajectory to drive the SDV towards the given target location based on the output from the perception and prediction modules \cite{fan2020computer}. The approaches designed under this paradigm are \textbf{non-interactive}, as the planning module is assumed to have no effects on the results of the prediction module. This implies that each SDV is a passive agent, and its behavior does not affect the other agents. Based on this philosophy, these non-interactive approaches typically perform acceptably in sparse traffic scenarios but can easily fail in dense traffic scenarios \cite{liu2021deep}. For instance, Fig.~\ref{fig.demo} (b) illustrates a dense traffic scenario where the non-interactive SDV tries to merge into the left lane. Since the estimated future trajectories of other agents can cover most of the road in such a dense scenario, it can be challenging for the SDV to plan a feasible trajectory to the left lane, resulting in a long wait for the SDV and causing inconvenience to the other agents behind it.

To address this problem, the SDV needs to be modeled as an active agent, capable of reasoning about how other agents will react to its behaviors. In this way, the prediction module becomes correlated to the planning module, and they can be formulated as a single \textbf{interactive} prediction and planning module, as shown  in Fig.~\ref{fig.demo} (c). An iterative SDV is now capable of considering the possible reactions of other agents when safely merging into the left lane in the same dense traffic scenario, as shown in Fig.~\ref{fig.demo}~(d). Researchers have already developed some interactive prediction and planning approaches from different perspectives, such as game-theoretic planning \cite{fisac2019hierarchical} and reinforcement learning \cite{saxena2020driving}. However, these approaches mainly depend on manually designed models or rewards, which may not be able to accurately model  real-world agent dynamics or human-like driving behaviors. Therefore, there is a strong motivation to develop a general interactive prediction and planning approach for autonomous driving.

In this paper, we propose an end-to-end \textbf{I}nteractive \textbf{N}eural \textbf{M}otion \textbf{P}lanner (INMP), for autonomous driving. Our INMP, illustrated in Fig.~\ref{fig.framework}, takes a set of past surrounding-view images and a high definition (HD) map as input to jointly 1) detect other agents and 2) perform interactive prediction and planning for the SDV. Specifically, we first lift these images into three dimensions (3-D), and then combine them with the HD map to generate a feature map in bird's-eye-view (BEV) space. This BEV feature map $\mathcal{F}_{b}$ is then processed to 1) detect other agents via a single-shot detection header and 2) jointly estimate the future trajectories of the detected agents and produce safe motion plans for the SDV via an interactive prediction and planning model with a joint probability distribution and a set of learnable costs. This paradigm enables the SDV to reason about how other agents will react to its behaviors, and thus can effectively improve the driving performance. Please note that the whole pipeline is differentiable, enabling end-to-end learning from raw sensor data to the outputs. In addition, we follow \cite{wang2021learning} and adopt a similar optical flow distillation paradigm to further improve the performance. Specifically, we refer to the above network as the student network (INMP-S) and additionally develop a teacher network (INMP-T), which adopts a similar architecture to the student network but further takes optical flow as an additional type of input. Optical flow can provide explicit motion information, leading to significant performance improvement for the teacher network. However, the computation of the optical flow seriously hinders the whole pipeline to achieve real-time performance \cite{sun2018pwc,wang2020cot}. We then distill the knowledge from the teacher network to the student network, which can effectively enhance the student network performance while still maintaining its real-time inference speed. We demonstrate the effectiveness and efficiency of our approach on the popular nuScenes dataset~\cite{caesar2020nuscenes} and in the closed-loop Carla simulation environment \cite{dosovitskiy2017carla}. Our INMP can achieve competitive performance on the detection, prediction, and planning tasks. Moreover, the adopted optical flow distillation paradigm enables our student network to achieve a much faster inference speed than the teacher network with similar driving performance.

\section{Related Work}
\label{sec.related_work}

\subsection{Trajectory Prediction}
Trajectory prediction aims to estimate the future trajectories of the agents based on their past states. The major challenges of this task are modeling the interactions between different agents and generating accurate multi-modal trajectory predictions. Traditional approaches generally achieve it based on manually designed models, \eg, the Kalman filter \cite{lefevre2014survey}. With the advancement of deep learning techniques, many data-driven approaches have achieved impressive performance in this field. These approaches typically use the past states to learn a latent representation for each agent, and model the interactions between different agents based on their latent representations \cite{alahi2016social,lee2017desire,gupta2018social}. Recently, some researchers have developed a new paradigm that takes the raw sensor data as input to jointly perform object detection and trajectory prediction \cite{zeng2019end,zeng2020dsdnet,liang2020pnpnet}. These approaches usually use LiDARs, since trajectory prediction is often performed in BEV space and the point clouds provided by LiDARs meet this requirement inherently. Considering that images can provide more semantic information than point clouds and cameras are much cheaper than LiDARs, we take images as input in our approach. Extensive experiments have demonstrated that the proposed vision-based approach can achieve competitive performance compared with previous LiDAR-based approaches, as presented in Section~\ref{sec.experiment}.

\subsection{Motion Planning}
The goal of motion planning is to generate a trajectory to drive the SDV towards its given destination safely and efficiently. Traditional approaches generally sample a large set of candidate trajectories based on the input perception and prediction results \cite{ozgunalp2016multiple,fan2020sne,fan2021learning,wang2020applying,wang2021dynamic}, and then use a cost function to select the executed trajectory, which has the minimal cost \cite{sadat2019jointly}. Recently, many end-to-end approaches that directly map the raw sensor data to the planned trajectories or control commands have been proposed \cite{gao2017intention,codevilla2018end,cai2020vtgnet}. These approaches are optimized jointly from data, and thus can compensate the adverse effects caused by the accumulated errors in traditional approaches \cite{levine2016end}. However, the end-to-end approaches are often criticized for their lack of interpretability, which makes these approaches hard to explain the generated behaviors and further leads to their limited applications in practice. To address it, some approaches have adopted the multi-task learning paradigm, jointly conducting detection, prediction, and planning tasks \cite{zeng2019end,zeng2020dsdnet,sadat2020perceive}. The generated intermediate results, \ie, detection and planning results, can effectively help people understand why the model can produce specific motion planning results. However, these approaches typically follow the non-interactive prediction and planning paradigm, and can easily fail in dense traffic scenarios, as mentioned above.

Recently, some researches have proposed the end-to-end interactive paradigm \cite{rhinehart2019precog,song2020pip,liu2021deep}. These approaches typically take the point clouds provided by LiDARs as input, and utilize joint probability distribution models to perform interactive prediction and planning. In this paper, we follow this paradigm and explore its feasibility and effectiveness when images are given as input.

\subsection{Knowledge Distillation}
Knowledge distillation aims at leveraging the dark knowledge of a teacher network to improve the performance of a student network with fewer parameters. This paradigm was first proposed in \cite{hinton2014distilling} for image classification. After that, researchers have presented  more effective and efficient knowledge distillation techniques \cite{romero2015fitnets,zagoruyko2017paying}. Specifically, \cite{romero2015fitnets} proposed hint training (HT), which aims at training the intermediate representation of the student network such that it can mimic the latent representation of the teacher network.  \cite{zagoruyko2017paying} defined attention maps for two networks and then forced the student network to mimic the attention maps of the teacher network. Knowledge distillation has been adopted in many other applications, \eg, object detection \cite{chen2017learning} and semantic segmentation~\cite{he2019knowledge}, to improve their performance. In this paper, we follow \cite{wang2021learning} and adopt a similar optical flow distillation paradigm for autonomous driving. Different from \cite{wang2021learning}, we utilize this technique for the detection, prediction, and planning tasks jointly. In addition, \cite{wang2021learning} adopts a non-interactive paradigm, while our INMP can perform interactive prediction and planning.

\section{Methodology}
\label{sec.methodology}
Fig.~\ref{fig.framework} illustrates the overview of the proposed approach. Our INMP first generates a BEV feature map $\mathcal{F}_{b}$, as introduced in Section~\ref{sec.bev_feature_map_generation}. $\mathcal{F}_{b}$ is then processed to 1) detect other agents and 2) perform interactive prediction and planning, as presented in Section~\ref{sec.object_detection} and Section~\ref{sec.interactive_prediction_and_planning}, respectively. After that, Section~\ref{sec.optical_flow_distillation_paradigm} elaborates the proposed optical flow distillation paradigm. Finally, we introduce the training phase in Section~\ref{sec.training_phase}.

\subsection{BEV Feature Map Generation}
\label{sec.bev_feature_map_generation}
\begin{figure}[t]
    \centering
    \includegraphics[width=0.99\linewidth]{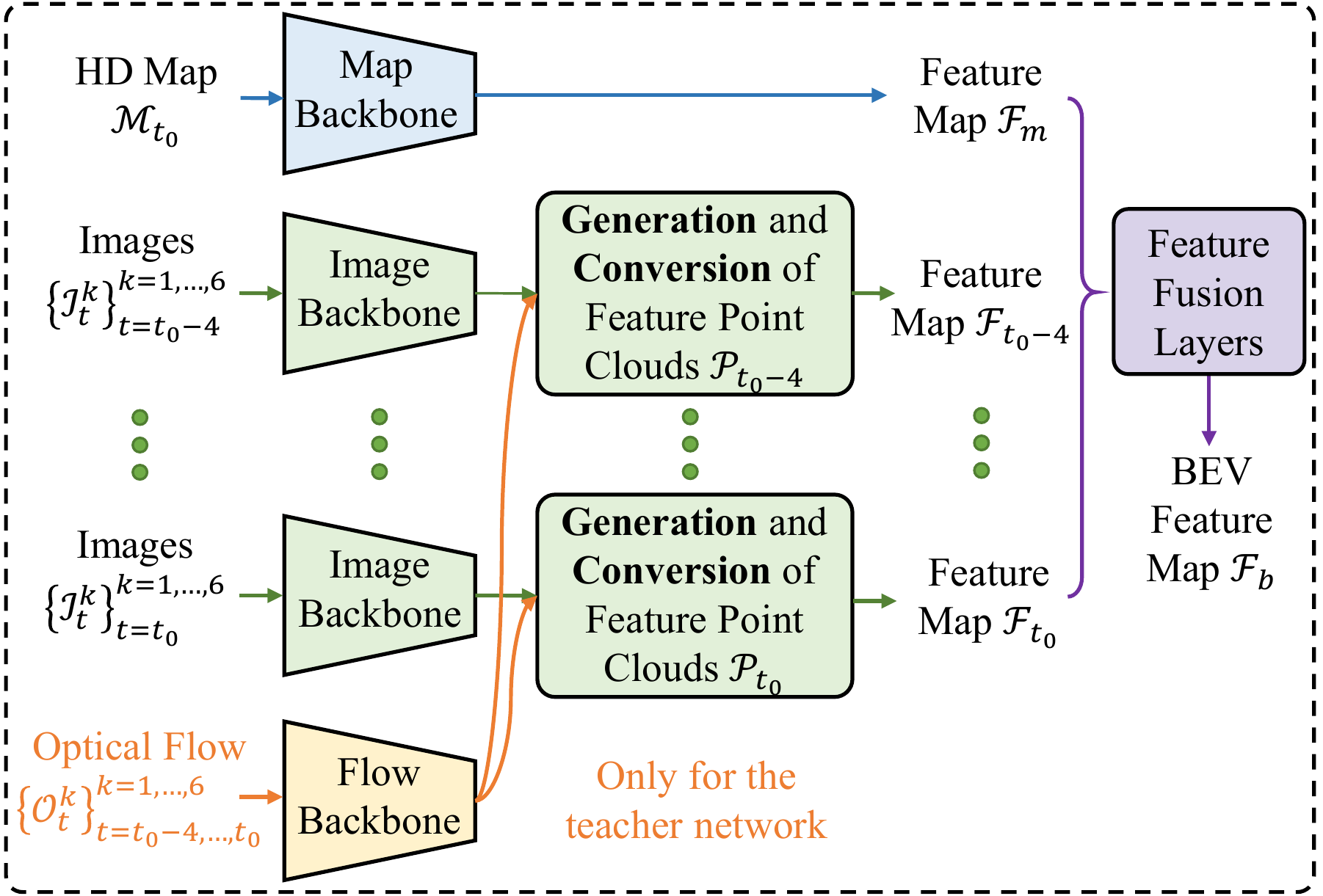}
    \caption{An illustration of the BEV feature map generation. We lift the input images into 3-D, and combine them with the HD map to generate a BEV feature map. The teacher network also uses optical flow to enhance its capability to model dynamic relationships for performance improvement.}
    \label{fig.module}
\end{figure}

Let $\mathcal{I}_{t}^{k} \in \mathbb{R}^{H \times W \times 3}$ denote the input RGB image, where $t = t_0-4, \dots, t_0$ denotes the timestamp of the past five frames; and $k = 1,\dots,6$ denotes the six cameras used in our experiments. The six cameras with known extrinsic and intrinsic parameters roughly point in the forward, forward-left, forward-right, backward, backward-left, and backward-right directions respectively. We also take the HD map $\mathcal{M}_{t_0}$ that contains the road, lane and intersection information as input, since it can provide a strong prior about the driving scenario. Then, given all images in the past five frames $\{\mathcal{I}_{t}^{k}\}_{t=t_0-4, \dots, t_0}^{k=1,\dots,6}$ and the current HD map $\mathcal{M}_{t_0}$, we aim to generate a BEV feature map $\mathcal{F}_{b}$, as presented in Fig.~\ref{fig.module}. $\mathcal{F}_{b}$ plays an important role in the following object detection and interactive prediction and planning.

Considering that images are located in perspective-view space, we first conduct monocular depth estimation for each $\mathcal{I}_{t}^{k}$, which builds a bridge between perspective-view space and BEV space. To achieve it, we follow \cite{philion2020lift} and generate contextual features at all possible depths for each pixel. Specifically, we associate each pixel with a set of $|\mathcal{D}|$ discrete depths, where $\mathcal{D} = \{d_0 + \Delta d, \dots, d_0 + |\mathcal{D}|\Delta d\}$. Then, we use the known intrinsic parameters to produce a point cloud $\mathcal{P}_t^k$ that contains $H \cdot W \cdot |\mathcal{D}|$ 3-D points for each $\mathcal{I}_{t}^{k}$. To obtain the contextual feature for each point in $\mathcal{P}_t^k$, we first use an image backbone to generate a contextual feature $\mathbf{f} \in \mathbb{R}^{C}$ and a distribution $\pi$ over the discrete depth set $\mathcal{D}$ for each pixel $\mathbf{p}$. Afterwards, the contextual feature $\mathbf{f}_d \in \mathbb{R}^{C}$ for point $\mathbf{p}_d$ is computed as a combination of the feature for the corresponding pixel and the discrete depth inference:
\begin{equation}
    \mathbf{f}_d = \pi_d \cdot \mathbf{f},
    \label{eq.contextual_feature}
\end{equation}
where $d \in \mathcal{D}$ refers to any discrete depth in $\mathcal{D}$.

For the teacher network, we incorporate optical flow information into $\mathcal{P}_t^k$ to enhance the network's capability to model dynamic relationships for performance improvement. Specifically, we use an existing optical flow estimation network \cite{sun2018pwc} to compute the backward optical flow $\mathcal{O}_{t}^{k} \in \mathbb{R}^{H \times W \times 2}$:
\begin{equation}
    \mathcal{I}_{t}^{k}(u, v) = \mathcal{I}_{t-1}^{k}\left(u + \mathcal{O}_{t}^{k}(u,v,1), v + \mathcal{O}_{t}^{k}(u,v,2)\right).
\end{equation}
$\mathcal{O}_{t}^{k}$ can be regarded as containing the explicit past motion information from $\mathcal{I}_{t-1}^{k}$ to $\mathcal{I}_{t}^{k}$. Then, we use a flow backbone to produce a contextual feature $\mathbf{f}' \in \mathbb{R}^{C}$ for each pixel $\mathbf{p}$. After that, we concatenate $\mathbf{f}'$ with $\mathbf{f}$ and produce a new feature. Please note that the new feature is still denoted as $\mathbf{f}$ for notational simplicity. However, $\mathbf{f}$ in the teacher network contains the explicit motion information provided by the optical flow while $\mathbf{f}$ in the student network does not. We then use \eqref{eq.contextual_feature} to compute a contextual feature $\mathbf{f}_{d} \in \mathbb{R}^{C}$ for every point $\mathbf{p}_d$ in the teacher network.

Then, we can use the known extrinsic parameters to aggregate $\{\mathcal{P}_{t}^{k}\}^{k=1,\dots,6}$ into a large point cloud $\mathcal{P}_{t}$ for each timestamp $t$. After that, we follow \cite{lang2019pointpillars} to convert $\mathcal{P}_{t}$ into ``pillars'', which refer to voxels with infinite height. To be specific, we assign each point to its nearest pillar and use pooling operation to construct a feature map $\mathcal{F}_{t}$, which contains the information in BEV space and can be processed by convolutional layers. We also use a map backbone to produce a feature map $\mathcal{F}_{m}$, which is then concatenated with the features of all five past frames $\{\mathcal{F}_{t}\}_{t=t_0, \dots, t=t_0-4}$ and processed by convolutional layers to generate the BEV feature map $\mathcal{F}_{b}$, as shown in Fig.~\ref{fig.module}.

\subsection{Object Detection}
\label{sec.object_detection}
Given the BEV feature map $\mathcal{F}_{b}$, we first detect other agents via a single-shot detection header. Specifically, following \cite{liu2016ssd}, we apply two convolutional layers on $\mathcal{F}_{b}$ separately, one for classifying the class of a location, the other one for regressing the position offset, size, heading angle, and velocity of each agent. After that, we use a non-maximum suppression (NMS) operation \cite{neubeck2006efficient} to obtain the bounding boxes and velocities of all other agents, which are then utilized to perform interactive prediction and planning.

The training loss for object detection $\mathcal{L}_{O}$ is defined as a summation of a classification loss $\mathcal{L}_{OC}$ and a regression loss $\mathcal{L}_{OR}$, \ie, $\mathcal{L}_{O} = \mathcal{L}_{OC} + \mathcal{L}_{OR}$. To be specific, in $\mathcal{L}_{OC}$, we use a cross entropy classification loss and assign the label of each anchor based on its intersection over union (IoU) with any agent as follows:
\begin{equation}
    \mathcal{L}_{OC} (\widehat{\mathcal{C}}, \mathcal{C}) = H\left( \widehat{\mathcal{C}}, \mathcal{C}\right),
    \label{eq.object_detection_classification}
\end{equation}
where $H(\cdot,\cdot)$ denotes the cross entropy; and $\widehat{\mathcal{C}}$ and $\mathcal{C}$ denote the ground-truth and the predicted classification distribution, respectively. In our experiments, we detect two kinds of agents, \ie, vehicles and pedestrians. In addition, $\mathcal{L}_{OR}$ is defined as a smooth $L_1$ loss between the regression ground truth $\widehat{\mathcal{S}}$ and regression predictions $\mathcal{S}$ as follows:
\begin{equation}
    \mathcal{L}_{OR} (\widehat{\mathcal{S}}, \mathcal{S}) = \sum_{k} SL_1 \left( \widehat{\mathcal{S}_k}, \mathcal{S}_k \right),
\end{equation}
where $SL_1(\cdot,\cdot)$ denotes the smooth $L_1$ loss; and the regression state set contains the position offsets in two dimensions, the width and height of the bounding box, the sine and cosine value of the orientation angle, and the velocities in two dimensions.

\subsection{Interactive Prediction and Planning}
\label{sec.interactive_prediction_and_planning}
Given the object detection results, we then focus on generating $\mathcal{T} = \left\{\tau_0, \tau_1, \dots, \tau_N \right\}$, which contains the planned trajectory of the SDV $\tau_0$ and the trajectory predictions of $N$ detected agents $\mathcal{T}_r = \left\{\tau_1, \dots, \tau_N \right\}$. Considering that performing prediction and planning in continuous space can consume much computational resources, we follow \cite{zeng2019end,zeng2020dsdnet,liu2021deep} and sample trajectories in a discrete space, which contains $K$ possible candidate trajectories for each trajectory $\tau \in \mathcal{T}$. The adopted trajectory sampler takes the past trajectories of each agent as input, and generates a set of straight lines, circular curves and euler spirals as candidate trajectories. Now, the generation of each trajectory $\tau \in \mathcal{T}$ is transformed to a classification problem.

To achieve it, we first define a joint probability distribution over the prediction and planning results $\mathcal{T}$ conditioned on the environmental context $\mathcal{X}$ as follows:
\begin{equation}
    p\left(\mathcal{T} | \mathcal{X}; \mathbf{w} \right) = \frac{1}{\mathcal{N}} \exp \left( -\mathcal{E}(\mathcal{T}, \mathcal{X}; \mathbf{w}) \right),
\end{equation}
where $\mathcal{N}$ is a normalizer; $\mathcal{X}$ includes the BEV feature map $\mathcal{F}_{b}$ and the past trajectories of each agent; $\mathcal{E}(\mathcal{T}, \mathcal{X}; \mathbf{w})$ denotes the defined joint energy of the prediction and planning results $\mathcal{T}$; and $\mathbf{w}$ denotes the parameters of the model. To be specific, $\mathcal{E}(\mathcal{T}, \mathcal{X}; \mathbf{w})$ is defined as a summation of an agent-specific term $\mathcal{E}_a$, a safety term $\mathcal{E}_s$ and a goal-directed term $\mathcal{E}_g$ as follows:
\begin{equation}
    \mathcal{E}(\mathcal{T}, \mathcal{X}; \mathbf{w}) = \sum_{i=0}^{N}\mathcal{E}_a(\tau_i, \mathcal{X}; \mathbf{w}) + \sum_{i,j}\mathcal{E}_s(\tau_i, \tau_j) + \mathcal{E}_g(\tau_0),
    \label{eq.energy}
\end{equation}
where $\mathcal{E}_a$ is used to evaluate all $K$ candidate trajectories for each agent; $\mathcal{E}_s$ is designed to penalize the occurrence of dangerous cases such as collision; and $\mathcal{E}_g$ is utilized to encourage the SDV to follow the input high-level route. Specifically, the BEV feature map $\mathcal{F}_{b}$ is combined with the candidate trajectories and then processed via a multi-layer perceptron (MLP) to produce a $K \times (N + 1)$ matrix of evaluation scores for $\mathcal{E}_a$. Moreover, we follow \cite{sadat2019jointly} and define $\mathcal{E}_s$ as a summation of a collision term and a safety distance violation term. The former will present 1 if the collision between a pair of future trajectories happens and 0 if not; while the latter is defined as a squared penalty within the safety distance of each agent's bounding box, scaled by the velocity of the SDV. Additionally, $\mathcal{E}_g$ is defined as the average projected distance between the planned trajectory of the SDV $\tau_0$ and the input high-level route.

Then, we can determine the planned trajectory of the SDV $\tau_0$ by selecting the candidate trajectory with the minimal cost of a defined cost function $f_I$ as follows:
\begin{equation}
    \tau_0^{*} = \mathop{\arg\min}_{\tau_0} f_I(\mathcal{T}, \mathcal{X}; \mathbf{w}).
\end{equation}
Compare with the non-interactive approaches, our INMP enables the SDV to reason about how other agents will react to its behaviors, \ie, considering the trajectory predictions of the other agents $\mathcal{T}_r$ conditioned on $\tau_0$ in the planning objective. Based on this philosophy, the cost function $f_I$ is defined as an expectation of the joint energy over the trajectory prediction distribution of the other agents conditioned on the planned trajectory of the SDV, as follows:
\begin{equation}
    f_I(\mathcal{T}, \mathcal{X} ; \mathbf{w})=\mathbb{E}_{\mathcal{T}_{r} \sim p \left(\mathcal{T}_{r} \mid \tau_0, \mathcal{X} ; \mathbf{w}\right)}\left[\mathcal{E} (\mathcal{T}, \mathcal{X} ; \mathbf{w})\right].
    \label{eq.cost}
\end{equation}
By substituting \eqref{eq.energy} into \eqref{eq.cost}, we can have
\begin{equation}
    \begin{aligned}
    f_I = \mathcal{E}_a(\tau_0) + \mathcal{E}_g(\tau_0) + \mathbb{E}_{\mathcal{T}_{r} \sim p \left(\mathcal{T}_{r} \mid \tau_0, \mathcal{X} ; \mathbf{w}\right)}[\sum_{i=1}^{N}\mathcal{E}_a(\tau_i)\\+\sum_{i=1}^{N}\mathcal{E}_s(\tau_0, \tau_i) +\sum_{i=1,j=1}^{N,N}\mathcal{E}_s(\tau_i, \tau_j)].
    \end{aligned}
    \label{eq.cost_exclude}
\end{equation}
where $f_I$ is short for $f_I(\mathcal{T}, \mathcal{X} ; \mathbf{w})$; and $\mathcal{E}_a(\tau_i)$ is short for $\mathcal{E}_a(\tau_i, \mathcal{X}; \mathbf{w})$. Since \cite{liu2021deep} have shown that excluding the interaction term between the other agents, \ie, $\sum_{i=1,j=1}^{N,N}\mathcal{E}_s(\tau_i, \tau_j)$, does not lead to significant performance degradation, we follow it and also exclude this term in \eqref{eq.cost_exclude} for computational efficiency. Then, for any given $\tau_0$, we can compute the expectation directly by simplifying the terms into the marginal probabilities as follows:
\begin{equation}
    \begin{aligned}
    f_I = \mathcal{E}_a(\tau_0) + \mathcal{E}_g(\tau_0) + \sum_{i,\tau_i}p\left(\tau_{i} \mid \tau_{0}, \mathcal{X} ; \mathbf{w}\right)\mathcal{E}_a(\tau_i)\\+\sum_{i,\tau_i}p\left(\tau_{i} \mid \tau_{0}, \mathcal{X} ; \mathbf{w}\right)\mathcal{E}_s(\tau_0, \tau_i).
    \end{aligned}
\end{equation}
Now, we can successfully perform the proposed interactive prediction and planning, \ie, estimating the future trajectories of the detected agents and producing safe motion plans for the SDV jointly, by computing these marginal probabilities via loopy belief propagation (LBP) \cite{yedidia2000generalized} efficiently.

Considering that the generation of each trajectory $\tau \in \mathcal{T}$ is transformed to a classification problem, we define the training loss for interactive prediction and planning $\mathcal{L}_{P}$ as follows:
\begin{equation}
    \mathcal{L}_{P}(\widehat{\mathcal{T}},\mathcal{T}) = \sum_{i} H\left( p(\widehat{\tau_i}), p(\tau_i)\right) + \sum_{i,j} H\left( p(\widehat{\tau_i}, \widehat{\tau_j}), p(\tau_i, \tau_j)\right),
    \label{eq.planning_loss}
\end{equation}
where $\widehat{\mathcal{T}} = \left\{\widehat{\tau_0}, \widehat{\tau_1}, \dots, \widehat{\tau_N} \right\}$ denotes the prediction and planning ground truth; and $p(\cdot)$ and $p(\cdot,\cdot)$ denote the marginal probabilities. Please note that we follow \cite{liu2021deep} and define $U(\widehat{\tau_i})$ as a set of the trajectories close to $\widehat{\tau_i}$. In \eqref{eq.planning_loss}, we only compute the loss for $\tau_i \notin U(\widehat{\tau_i})$, since we do not want to penalize the trajectory close to the ground truth.

\subsection{Optical Flow Distillation Paradigm}
\label{sec.optical_flow_distillation_paradigm}
As mentioned previously, optical flow can provide explicit motion information, leading to significant performance improvement for the teacher network. However, the computation of the optical flow seriously hinders the whole pipeline to achieve real-time performance. We then follow~\cite{wang2021learning} and distill the knowledge from the teacher network to the student network, which can effectively enhance the student network while still maintaining its real-time performance. The distillation loss $\mathcal{L}_{D}$ is defined as follows:
\begin{equation}
    \mathcal{L}_{D} = \lambda_{DO} \mathcal{L}_{DO} + \lambda_{DP} \mathcal{L}_{DP} + \lambda_{DF} \mathcal{L}_{DF},
\end{equation}
where $\mathcal{L}_{DO}$, $\mathcal{L}_{DP}$ and $\mathcal{L}_{DF}$ denote the distillation loss for object detection, interactive prediction and planning, and the BEV feature map $\mathcal{F}_b$, respectively; and $\lambda_{DO}$, $\lambda_{DP}$ and $\lambda_{DF}$ are the hyperparameters that scale the three loss terms.

Similar to $\mathcal{L}_{O}$, $\mathcal{L}_{DO}$ is defined as a summation of a classification distillation loss $\mathcal{L}_{DOC}$ and a regression loss $\mathcal{L}_{DOR}$, \ie, $\mathcal{L}_{DO} = \mathcal{L}_{DOC} + \mathcal{L}_{DOR}$. We follow \cite{hinton2014distilling} and define $\mathcal{L}_{DOC}$ as follows:
\begin{equation}
    \mathcal{L}_{DOC} = \mathcal{L}_{OC} (\mathcal{C}^T, \mathcal{C}^S) = H\left( \mathcal{C}^T, \mathcal{C}^S\right),
\end{equation}
where $\mathcal{C}^T$ and $\mathcal{C}^S$ denote the predicted classification distributions of the teacher and student networks, respectively. Different from $\widehat{\mathcal{C}}$ in \eqref{eq.object_detection_classification} that can only provide hard information, $\mathcal{C}^{T}$ can provide useful soft information to effectively improve the student network. In addition, inspired by \cite{chen2017learning}, we design $\mathcal{L}_{DOR}$ as follows:
\begin{equation}
    \small
    \mathcal{L}_{DOR}=\sum_{k}\left\{
        \begin{array}{ll}
         SL_1 \left(\mathcal{S}_k^T, \mathcal{S}_k^S \right), \text {if} ||\widehat{\mathcal{S}_k} - \mathcal{S}_k^S||_{1} > ||\widehat{\mathcal{S}_k} - \mathcal{S}_k^T||_{1}, \\
        0, \hspace{1.8cm}\text {otherwise.}
    \end{array}\right.
\end{equation}
where $||\cdot||_1$ denotes the $L_1$ norm; and $\mathcal{S}^T$ and $\mathcal{S}^S$ denote the regression predictions of the teacher and student networks, respectively. $\mathcal{L}_{DOR}$ encourages the student network to be close or better than the teacher network, but does not push the student once it reaches the teacher's performance.

Moreover, we define $\mathcal{L}_{DP}$ as follows:
\begin{equation}
    \mathcal{L}_{DP}=\left\{
        \begin{array}{ll}
        \mathcal{L}_{P}(\mathcal{T}^T,\mathcal{T}^S), \text{if} \sum_i D(\widehat{\tau_i}, \tau_i^{S*}) > \sum_i D(\widehat{\tau_i}, \tau_i^{T*}), \\
        0, \hspace{1.6cm}\text {otherwise.}
    \end{array}\right.
\end{equation}
where $\mathcal{T}^T$ and $\mathcal{T}^S$ denote the prediction and planning results of the teacher and student networks, respectively; $\widehat{\tau_i}$ denotes the trajectory ground truth; $\tau_i^{T*}$ and $\tau_i^{S*}$ denote the trajectories of the teacher and student networks with the minimal cost of $f_I$, respectively; and $D(\cdot,\cdot)$ measures the average projected distance between two trajectories. Similar to $\mathcal{L}_{DOR}$, $\mathcal{L}_{DP}$ also encourages the student network to perform better than the teacher, network but does not push the student too much.

Considering that $\mathcal{F}_b$ of the teacher network incorporates the explicit motion information provided by the optical flow while $\mathcal{F}_b$ of the student network does not, we follow HT~\cite{romero2015fitnets} and further design $\mathcal{L}_{DF}$ as:
\begin{equation}
    \mathcal{L}_{DF} = ||\mathcal{F}_{b}^{T}-\mathcal{F}_{b}^{S}||_{1},
\end{equation}
where $\mathcal{F}_{b}^{T}$ and $\mathcal{F}_{b}^{S}$ denote the BEV feature maps $\mathcal{F}_{b}$ of the teacher and student networks, respectively. $\mathcal{L}_{DF}$ encourages the student network to mimic the BEV feature map $\mathcal{F}_{b}$ of the teacher network.

\subsection{Training Phase}
\label{sec.training_phase}
In the training phase, we first use the following teacher training loss $\mathcal{L}^{T}$ to train the teacher network:
\begin{equation}
    \mathcal{L}^{T} = \lambda_{O} \mathcal{L}_{O}^{T} + \lambda_{P} \mathcal{L}_{P}^{T},
\end{equation}
where $\mathcal{L}_{O}^{T} = \mathcal{L}_{OC} (\widehat{\mathcal{C}}, \mathcal{C}^T) + \mathcal{L}_{OR} (\widehat{\mathcal{S}}, \mathcal{S}^T)$; $\mathcal{L}_{P}^{T} = \mathcal{L}_{P} (\widehat{\mathcal{T}}, \mathcal{T}^T)$; and $\lambda_{O}$ and $\lambda_{P}$ are the hyperparameters that scale the two loss terms.

After that, we utilize the following student training loss $\mathcal{L}^{S}$ to train the student network based on the trained teacher network:
\begin{equation}
    \mathcal{L}^{S} = \lambda_{O} \mathcal{L}_{O}^{S} + \lambda_{P} \mathcal{L}_{P}^{S} + \lambda_{D} \mathcal{L}_{D},
\end{equation}
where $\mathcal{L}_{O}^{S} = \mathcal{L}_{OC} (\widehat{\mathcal{C}}, \mathcal{C}^S) + \mathcal{L}_{OR} (\widehat{\mathcal{S}}, \mathcal{S}^S)$; $\mathcal{L}_{P}^{T} = \mathcal{L}_{P} (\widehat{\mathcal{T}}, \mathcal{T}^S)$; and $\lambda_{O}$, $\lambda_{P}$ and $\lambda_{D}$ are the hyperparameters that scale the three loss terms.

\begin{figure*}[t]
    \centering
    \includegraphics[width=0.99\textwidth]{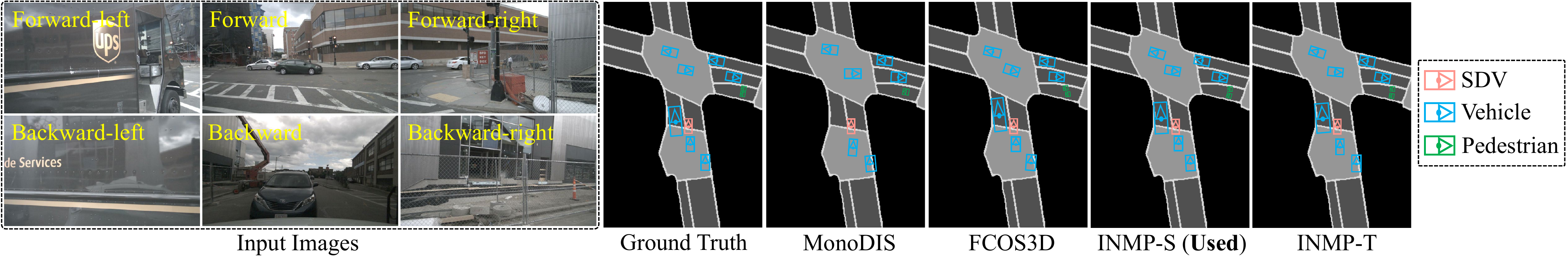}
    \caption{Object detection results of MonoDIS \cite{simonelli2019disentangling}, FCOS3D \cite{mmdet3d2020}, our INMP-S, and our INMP-T on the nuScenes dataset \cite{caesar2020nuscenes}. It is evident that our INMP-T and INMP-S can produce more accurate results than other approaches.}
    \label{fig.detection}
\end{figure*}

\section{Experimental Results and Discussions}
\label{sec.experiment}

\subsection{Datasets and Implementation Details}
\label{sec.datasets_and_implementation_details}
In our experiments, we first evaluate the performance of our approach for object detection and trajectory prediction on the nuScenes dataset \cite{caesar2020nuscenes}, which contains around 1000 human driving sequences. The dataset is split into a training, a validation and a test set that consists of 18072, 8019 and 8033 samples, respectively. The best-performing networks are selected on the validation set and evaluated on the test set. We also conduct closed-loop evaluation in the Carla simulation environment \cite{dosovitskiy2017carla}. Specifically, we first collect a large-scale driving dataset on different maps with different weather and illumination conditions, \eg, clear, rainy, daytime and sunset. We also set random roaming pedestrians and vehicles, which are controlled by the Carla simulator \cite{dosovitskiy2017carla}. The dataset is split into a training set with 200K samples and a validation set with 50K samples. Finally, each network is evaluated thoroughly with 1800 episodes (around $1000km$) in a closed-loop manner.

\begin{table}[t]
    \centering
    \caption{Object detection results (\%) on the nuScenes dataset \cite{caesar2020nuscenes}. Best results are shown in the bold type.}
    \begin{tabular}{L{2.3cm}C{1.6cm}C{1.9cm}C{0.8cm}}
        \toprule
        Approach & $\text{AP}_{vehicle}$ & $\text{AP}_{pedestrian}$ & mAP \\ \midrule
        MonoDIS \cite{simonelli2019disentangling} & 45.67 & 36.70 & 41.19 \\
        CenterNet \cite{duan2019centernet} & 52.06 & 37.85 & 44.96 \\
        FCOS3D \cite{mmdet3d2020} & 51.34 & 39.21 & 45.28 \\ \midrule
        INMP-S-ND & 49.79 & 38.95 & 44.37 \\
        INMP-S (\textbf{Used}) & 52.58 & \textbf{40.63} & 46.61 \\
        INMP-T & \textbf{53.81} & 40.34 & \textbf{47.08} \\
        \bottomrule
    \end{tabular}
    \label{tab.eval_detection}
\end{table}

For the implementation details, we adopt EfficientNet-B0 \cite{tan2019efficientnet} as the map, flow and image backbones. In addition, our INMP takes the information of past $2s$ as input and performs interactive prediction and planning for the future $4s$. We adopt the Adam optimizer \cite{kingma2014adam} with an initial learning rate of $10^{-4}$ to train our INMP-T and INMP-S on two NVIDIA GeForce RTX 2080 Ti GPUs. Moreover, we train the student network without the proposed optical flow distillation paradigm, referred to as INMP-S-ND, for performance comparison.

\subsection{Object Detection Results}
\label{sec.object_detection_results}
We follow the nuScenes benchmark \cite{caesar2020nuscenes} and adopt the average precision (AP) at the $1m$ distance threshold as our evaluation metric. We compute the AP for vehicles and pedestrians respectively, and also compute its mean value (mAP) across the two classes. The evaluation results are shown in Table~\ref{tab.eval_detection}. It is evident that the three variants of our INMP all achieve competitive performance compared to the existing vision-based approaches, and our INMP-T achieves the best performance. In addition, our INMP-S presents a better performance than INMP-S-ND and a similar performance to INMP-T thanks to the adopted optical flow distillation paradigm. The qualitative results in Fig.~\ref{fig.detection} also confirm the above conclusions. Moreover, we adopt INMP-S in practice due to its real-time inference speed, as analyzed in Section~\ref{sec.eval_closed_loop}.

\begin{table}[t]
    \centering
    \caption{Trajectory prediction results ($m$) on the nuScenes dataset~\cite{caesar2020nuscenes}. Best results are shown in the bold type.}
    \begin{tabular}{L{2.3cm}C{2.0cm}C{1.0cm}C{1.3cm}}
        \toprule
        Approach & Type & $L_2$ & minMSD \\ \midrule
        NMP \cite{zeng2019end} & LiDAR-based & 2.36 & 3.22 \\
        ESP \cite{rhinehart2019precog} & LiDAR-based & 2.15 & 2.93 \\
        DSDNet \cite{zeng2020dsdnet} & LiDAR-based & 2.04 & 2.65 \\ \midrule
        INMP-S-ND & Vision-based & 2.29 & 3.10 \\
        INMP-S (\textbf{Used}) & Vision-based & 2.07 & 2.68 \\
        INMP-T & Vision-based & \textbf{1.95} & \textbf{2.59} \\
        \bottomrule
    \end{tabular}
    \label{tab.eval_prediction}
\end{table}

\subsection{Trajectory Prediction Results}
\label{sec.trajectory_prediction_results}
We use the $L_2$ distance at $t=4s$ \cite{zeng2020dsdnet} and the minMSD (5 agents and $K=12$) \cite{rhinehart2019precog} for performance comparison between trajectory prediction approaches. These two metrics both measure the distance between the trajectory prediction and the ground truth for the correctly detected agents, and the evaluation results are presented in Table~\ref{tab.eval_prediction}. We can see that the conclusions in Section~\ref{sec.object_detection_results} also hold for the trajectory prediction task. Excitingly, our INMP-S and INMP-T can even achieve competitive performance compared to existing LiDAR-based approaches, which strongly demonstrates that our energy-based model can effectively generate accurate trajectory predictions.

\subsection{Closed-loop Evaluation Results}
\label{sec.eval_closed_loop}
We adopt the success rate (SR) and right lane rate (RL) as our evaluation metrics \cite{cai2020probabilistic}. SR is defined as the proportion of the successfully finished episodes to the total testing episodes, while RL is defined as the proportion of the period when the SDV drives in the input high-level route to the total driving time. To verify the effectiveness of our INMP, we further develop a non-interactive neural motion planner (NINMP) by using the following cost function: $f_{NI}(\mathcal{T}, \mathcal{X} ; \mathbf{w})=\mathbb{E}_{\mathcal{T}_{r} \sim p \left(\mathcal{T}_{r} \mid \mathcal{X} ; \mathbf{w}\right)}\left[C(\mathcal{T}, \mathcal{X} ; \mathbf{w})\right]$. Different from \eqref{eq.cost}, $f_{NI}$ considers the trajectory predictions of the other agents $\mathcal{T}_r$ unconditioned on the planned trajectory of the SDV $\tau_0$. Moreover, we record the inference time of each approach on the NVIDIA GeForce RTX 2080 Ti GPU.

\begin{table}[t]
    \centering
    \caption{Closed-loop evaluation results in the Carla simulator~\cite{dosovitskiy2017carla}. Best results are shown in the bold type.}
    \begin{tabular}{L{2.5cm}C{1.3cm}C{1.3cm}C{1.3cm}}
        \toprule
        Approach & SR~(\%) & RL~(\%) & Time~($s$) \\ \midrule
        CIL \cite{codevilla2018end} & 60.72 & 82.97 & 0.07 \\
        VTP \cite{cai2019vision} & 76.11 & 80.89 & 0.08 \\ \midrule
        NINMP-S & 80.94 & 82.31 & \textbf{0.05} \\
        INMP-S-ND & 81.39 & 85.63 & 0.06 \\
        INMP-S (\textbf{Used}) & 91.28 & 93.96 & 0.06 \\
        INMP-T & \textbf{92.33} & \textbf{95.20} & 0.21 \\
        \bottomrule
    \end{tabular}
    \label{tab.eval_carla}
\end{table}

\begin{table}[t]
    \centering
    \caption{Closed-loop evaluation results of our INMP-S with some of the loss terms disabled in the Carla simulator~\cite{dosovitskiy2017carla}. Best results are shown in the bold type.}
    \begin{tabular}{L{2.8cm}C{0.75cm}C{0.75cm}C{0.75cm}C{1.15cm}}
        \toprule
        Variant & $\mathcal{L}_{DO}$ & $\mathcal{L}_{DP}$ & $\mathcal{L}_{DF}$ & SR~(\%) \\ \midrule
        (a) INMP-S & -- & -- & -- & 81.39 \\
        (b) INMP-S & \cmark & \cmark & -- & 89.61 \\
        (c) INMP-S & \cmark & -- & \cmark & 83.39 \\
        (d) INMP-S & -- & \cmark & \cmark & 87.83 \\
        (e) INMP-S (\textbf{Used}) & \cmark & \cmark & \cmark & \textbf{91.28} \\
        \bottomrule
    \end{tabular}
    \label{tab.ablation_study}
\end{table}

\begin{figure}[t]
    \centering
    \includegraphics[width=0.99\linewidth]{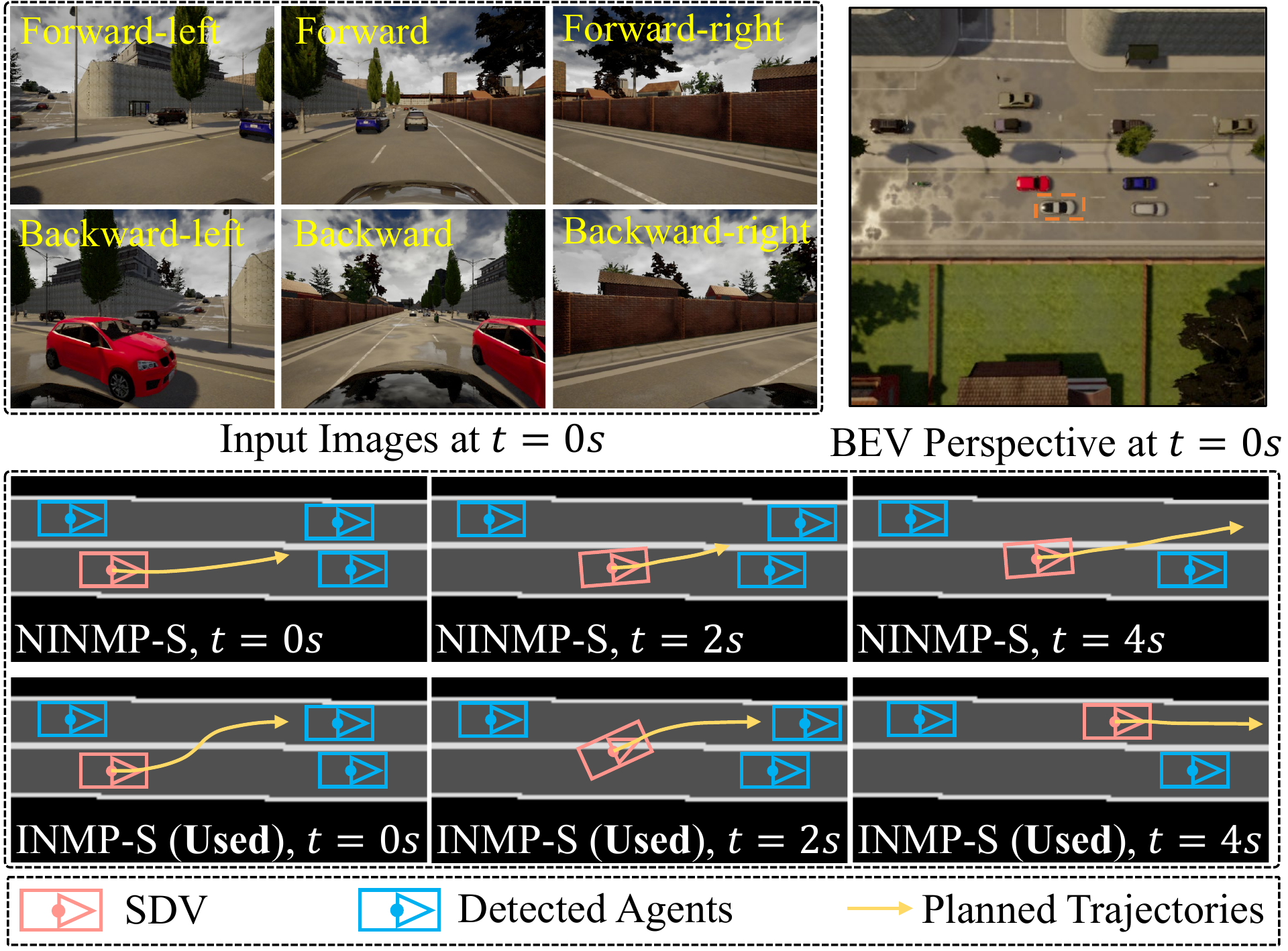}
    \caption{An example scenario in the Carla simulator \cite{dosovitskiy2017carla}, where the SDV is trying to merge into the left lane. The non-interactive SDV (NINMP-S) drifts slowly to the left lane instead of completing a lane merge, while our interactive SDV (INMP-S) can complete a satisfactory lane merge efficiently without bringing safety risk and inconvenience to other agents. The SDV is marked with an orange dashed box in the BEV perspective.}
    \label{fig.carla}
\end{figure}

Table~\ref{tab.eval_carla} presents the evaluation results, where it is evident that the conclusions in Section~\ref{sec.object_detection_results} also hold for the closed-loop autonomous driving task. Our INMP-T achieves the best performance thanks to the explicit motion information provided by the optical flow. Moreover, our INMP-S can present a real-time inference speed with a similar performance to INMP-T, which demonstrates the effectiveness of our optical flow distillation paradigm. This is also the reason why we adopt INMP-S in practice. In addition, our INMP-S outperforms the non-interactive approach, NINMP-S, in terms of both SR and RL. This implies that by considering how other agents will react to the SDV's behaviors in the planning objective, our interactive prediction and planning model can effectively improve the driving performance for the SDV. Fig.~\ref{fig.carla} presents an example scenario, where the SDV is trying to merge into the left lane. It is evident that the non-interactive SDV (NINMP-S) drifts slowly to the left lane instead of completing a lane merge, while our interactive SDV (INMP-S) can complete a satisfactory lane merge efficiently without bringing safety risk and inconvenience to other agents. All the analysis have demonstrated the effectiveness and efficiency of our proposed approach.

\subsection{Ablation Study}
\label{sec.ablation_study}
We conduct ablation studies to demonstrate the effectiveness of our selection in the loss functions. Specifically, we take INMP-S as the baseline, and test its closed-loop performance with different combinations of loss terms in the Carla simulator \cite{dosovitskiy2017carla}. The evaluation results are presented in Table~\ref{tab.ablation_study}, where it can be observed that our paradigm that employs the three loss terms together achieves the best performance for our INMP-S.

\section{Conclusions}
\label{sec.conclusions}
In this paper, we proposed INMP, an end-to-end interactive neural motion planner for autonomous driving. Given a set of past surrounding-view images and a high definition map, our INMP first generated a feature map in bird's-eye-view space, which was then processed to detect other agents and perform interactive prediction and planning jointly. Our interactive prediction and planning paradigm enables the self-driving vehicle to reason about how other agents will react to its behaviors, and thus can significantly improve the driving performance. In addition, we adopted an optical flow distillation paradigm, which can effectively improve the network performance while still maintaining its real-time inference speed. Extensive experiments on the nuScenes dataset and in the closed-loop Carla simulation environment demonstrated the effectiveness and efficiency of our INMP for the detection, prediction, and planning tasks. In the future, we plan to utilize the optical flow distillation paradigm in other tasks related to spatio-temporal information analysis for their performance improvement.

\clearpage

{\small
\bibliographystyle{ieee_fullname}
\bibliography{egbib}
}

\end{document}